\documentclass{article}

\usepackage{arxiv}

\usepackage[utf8]{inputenc} 
\usepackage[T1]{fontenc}    
\usepackage{hyperref}       
\usepackage{url}            
\usepackage{booktabs}       
\usepackage{amsfonts}       
\usepackage{nicefrac}       
\usepackage{microtype}      
\usepackage{graphicx}
\usepackage{natbib}
\usepackage{doi}
\usepackage{lmodern}
\usepackage{multirow}
\usepackage{array}
\usepackage{xcolor}
\usepackage{colortbl}
\usepackage{makecell}
\usepackage{siunitx}
\usepackage{enumitem}   
\usepackage{amsmath}    
\usepackage{cleveref}
\usepackage{tikz}
\usetikzlibrary{shapes.geometric,arrows.meta,positioning}

\newcommand{\best}[1]{\textbf{#1}}

\title{Efficient Multilingual Neural Machine Translation via Corpus-Driven Vocabulary Pruning: An English-Arabic Case Study}

\usepackage{authblk}

\author[1,3]{Ahmed Amine Aliane\thanks{Corresponding author: \texttt{aaliane781@gmail.com}}}
\author[2]{Nasredine Semmar}
\author[3]{Hassina Aliane}

\affil[1]{Arabic Institute for Translation, Algiers, Algeria}
\affil[2]{CEA LIST, Paris-Saclay, France}
\affil[3]{CERIST, Algeria}

\renewcommand{\shorttitle}{Efficient Multilingual NMT via Vocabulary Pruning}

\hypersetup{
	pdftitle={A General Framework for Efficient Multilingual Neural Machine Translation},
	pdfsubject={cs.CL, cs.LG},
	pdfauthor={Ahmed Amine Aliane, Nasredine Semmar, Hassina Aliane},
	pdfkeywords={Neural Machine Translation, Vocabulary Pruning, Arabic NLP, Multilingual Models},
}

\begin{document}
	\maketitle

	\begin{abstract}
		The adoption of large pre-trained multilingual models for neural machine translation
		(MNMT) faces a major challenge: excessive memory and computational consumption due to
		overly large vocabularies and embedding layers. Although existing compression methods like
		pruning, quantization and knowledge distillation reduce parameter redundancy, they mainly
		preserve the structure of the original vocabulary, thereby leaving a major source of
		inefficiency unresolved. We propose in this paper a general optimization framework that
		combines a vocabulary pruning method with a targeted fine-tuning protocol for MNMT
		models. We evaluate the proposed framework using three models (M2M100, NLLB-200,
		mBART-50) on the English-Arabic language pair. Our approach reduces the vocabulary size
		from over 128,000 to approximately 10,000 tokens, enabling a 60\% memory saving without
		any loss in performance.
		Results show that optimized multilingual models can match or exceed the performance of
		dedicated bilingual baselines. In particular, the pruned and fine-tuned M2M100 model
		achieves a competitive BLEU score of 42.04 (against 44.59 for the OPUS-MT-en-ar bilingual
		model) while it significantly outperforms it on the COMET metric (0.8730 vs 0.7911) revealing
		superior semantic adequacy and fluency.
	\end{abstract}


	\section{Introduction}
	Large-scale Multilingual Neural Machine Translation models such as Facebook's M2M100
	and Meta's NLLB-200 have transformed cross-lingual interaction by facilitating translation
	among hundreds of language combinations through a single architecture. These models have
	achieved state-of-the-art performance and significantly advanced zero-shot translation
	capabilities. However, to achieve high performance, they rely on massive parameter counts
	and require very large vocabularies (often exceeding 128,000 subword tokens), resulting in
	prohibitive memory footprints and inference latency. For many researchers working in
	resource-constrained environments, the enormous size of the embedding and output projection
	layers poses a considerable challenge. Indeed, for a specific language pair, most of the more
	than 128,000 vocabulary entries remain inactive. While existing compression techniques such
	as quantization, and weight pruning have been extensively studied,
	they primarily focus on reducing model parameters rather than addressing the structural
	inefficiency introduced by large multilingual vocabularies. Moreover, most adaptation
	methods preserve the original vocabulary, thereby maintaining a high memory footprint even
	after compression. In practice, very few studies explicitly target the vocabulary size issue in
	large-scale MNMT systems, leaving an important gap in efficient multilingual model design.

	In this paper, we propose a novel general optimization framework that combines vocabulary
	pruning and targeted fine-tuning for multilingual neural machine translation. First, our
	vocabulary pruning method effectively reduces the vocabulary of the pretrained multilingual
	models, retaining only the tokens relevant to a specific language pair, which considerably
	diminishes the size of the embeddings matrices and eliminates the noise of unused languages.
	Then, we introduce a targeted fine-tuning protocol that adapts the pruned model to the
	specific translation task to allow the model to re-learn the optimal parameters configuration
	within the new vocabulary space.

	We evaluate the proposed framework using three state-of-the-art multilingual models
	(M2M100, mBART-50, NLLB-200) for English-Arabic machine translation and the dedicated
	model OPUS-MT-en-ar as a baseline.
	Furthermore, we trained our models on both formal and hybrid (formal and informal) corpora,
	to explore the impact of data composition on the stability of pruned models. Our contributions
	are summarized as follows:
	\begin{enumerate}
		\item Unlike other compression methods which often require architectural modification or
		extensive retraining, we introduce a minimalist architecture-agnostic framework for MNMT
		that is founded on vocabulary pruning.
		\item We validate the proposed framework through extensive experiments on the English-Arabic
		translation task using M2M100, mBART-50 and NLLB-200 obtaining scores comparable or exceeding those obtained with a dedicated bilingual model, with a reduction
		of 60\% of memory footprint.
		\item We analyze the synergy between vocabulary reduction and fine-tuning: results show that
		hybrid data strategies combining formal and informal text are crucial for maintaining model
		robustness after compression, outperforming models trained on single-register corpora.
	\end{enumerate}

	\section{Related work}
	Our work lies at the intersection of MNMT, model compression and domain adaptation. We
	present in this section an overview of key advances in these domains.

	\subsection{Large-scale Multilingual Neural Machine Translation Models}
	Multilingual Neural Machine Translation (MNMT) is a relatively recent approach built on the
	NMT successes. MNMT's key innovation rests on leveraging a single shared model capable
	of translating between multiple languages instead of training a specific model for each
	targeted language pair \citep{firat2016multiway, johnson2017google}. The adoption of subword
	tokenization \citep{kudo2018subword}, which provided an effective means of addressing the rare
	word problem in neural machine translation \citep{Luong2015addressing}, has subsequently
	become an essential component of multilingual models by enabling a shared vocabulary across
	diverse languages and writing systems. The introduction of mBART
	\citep{liu2020multilingual} marked an important milestone, demonstrating the effectiveness of
	multilingual denoising pretraining for sequence-to-sequence translation for 25 languages.
	mBART-50 \citep{tang2020multilingual} extended this capability to a many-to-many setting, handling
	50 languages without relying on English as an interlingua. M2M100 \citep{fan2021beyond} scaled
	multilingual translation to 100 languages through a natively trained many-to-many
	architecture while NLLB-200 \citep{nllb2022no} expanded this capability to 200
	languages, achieving state-of-the-art results across diverse language families. Nevertheless,
	although these models have significantly advanced the quality and language coverage of
	multilingual machine translation, they come at the cost of prohibitive memory requirements
	and computational overhead.

	\subsection{Model compression and domain adaptation}
	Extensive research has explored the use of compression techniques and adaptation strategies
	to improve the efficiency of large-scale multilingual neural translation models. Common
	approaches, including quantization \citep{zafrir2019q8bert, shen2020qbert, mohammadshahi2022what, prato2020fully},
	weight pruning \citep{see2016compression, voita2019analyzing, behnke2020losing, khan2021more},
	layer pruning \citep{fan2019layer, peer2022greedy, moslem2025layer}
	and knowledge distillation \citep{hinton2015distilling}, \citep{kim2016sequence}, \citep{tan2019multilingual} have
	been successfully applied to monolingual and multilingual models. Recent studies have shown
	that compressing multilingual models is not language-neutral, with pruning and quantization
	potentially causing uneven degradation across languages, particularly for low-resource languages
	\citep{mohammadshahi2022what, gaido2025findings, palomino2026selected}. However, these methods
	primarily compress the model's parameters while leaving the embedding layers largely
	unchanged, which constitutes a structural bottleneck for multilingual models where the shared
	vocabulary size often exceeds 128,000 tokens, contributing substantially to the memory
	footprint.

	Domain adaptation via targeted fine-tuning is a well-established strategy for specializing
	multilingual models to specific language pairs, yet it typically preserves the full multilingual
	vocabulary. Vocabulary pruning, the direct reduction of the model's vocabulary by
	removing tokens irrelevant to a target language pair, is a less explored avenue in MNMT
	literature. \citet{lhostis2016vocabulary} early explored vocabulary selection and pruning strategies for efficient
	NMT, in bilingual settings, but did not address the challenges introduced by large-scale
	multilingual models and shared cross-lingual vocabularies. Existing work has mainly investigated
	vocabulary trimming \citep{ushio-etal-2023-efficient, dorkin2024tokenization},
	tokenizer adaptation and re-tokenization strategies, and the impact of
	BPE vocabulary trimming on translation quality \citep{cognetta2024analysis}. More recently,
	\citet{jiang2026tokenizer} proposed a method based on re-tokenizing a pruned vocabulary and
	relearning embeddings; the method significantly improved performance on low-resource and
	non-Latin-script languages. However, despite
	these efforts, combining domain adaptation with vocabulary reduction for pretrained
	multilingual translation models remains underexplored.

	\section{Methodology}
	\label{sec:proposed}

	Unlike common compression methods, our framework employs a deterministic pruning
	strategy that preserves the original weight alignment for non-embedding layers while
	systematically removing only the tokens irrelevant to the target language pair.

	Our approach is motivated by two main observations. First, conventional compression
	methods primarily target the dense weight matrices of the attention and feed-forward layers
	while leaving the vocabulary-dependent layer largely unchanged. Consequently, a substantial
	source of memory overhead remains unaddressed: the input embedding and output projection
	matrices whose size linearly grows with the multilingual vocabulary \citep{fan2021beyond}.
	However, for language-pair specific translation tasks, only a small portion of this shared
	vocabulary is actively utilized while the remaining embedding parameters continue to occupy
	memory throughout both fine-tuning and inference. This argues for vocabulary pruning as an
	orthogonal direction for compressing multilingual translation models. Second, large
	multilingual models such as M2M100 and NLLB-200 are jointly pretrained on hundreds of
	languages and billions of parallel sentences. This large-scale multilingual training enables the
	model to learn language-agnostic representations that transfer effectively across language
	pairs, providing a substantially stronger initialization for bilingual fine-tuning than training a
	bilingual model from scratch \citep{arivazhagan2019massively}. By modeling semantic relations
	across diverse languages, scripts and linguistic structures, the shared encoder captures cross-
	lingual regularities that are difficult to acquire from bilingual data alone. This observation
	supports the hypothesis that the bulk of the multilingual knowledge acquired during
	pretraining resides within the structural shared parameters rather than being strictly bound to
	the exhaustive joint vocabulary.

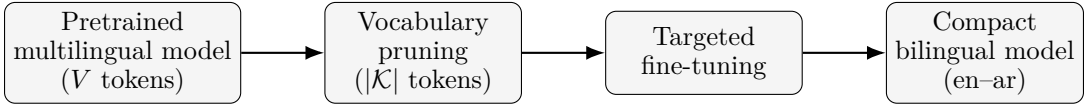
\begin{figure}[h]
		\centering
		\begin{tikzpicture}[
			node distance=0.9cm and 1.1cm,
			box/.style={rectangle, draw, rounded corners, minimum height=1.1cm, minimum width=2.6cm, align=center, fill=gray!8},
			arr/.style={-{Latex[length=2.5mm]}, thick}
			]
			\node[box] (pre) {Pretrained\\multilingual model\\($V$ tokens)};
			\node[box, right=of pre] (prune) {Vocabulary\\pruning\\($|\mathcal{K}|$ tokens)};
			\node[box, right=of prune] (ft) {Targeted\\fine-tuning};
			\node[box, right=of ft] (out) {Compact\\bilingual model\\(en--ar)};
			\draw[arr] (pre) -- (prune);
			\draw[arr] (prune) -- (ft);
			\draw[arr] (ft) -- (out);
		\end{tikzpicture}
		\caption{Overview of the proposed optimization framework: corpus-driven vocabulary pruning followed by targeted fine-tuning.}
		\label{fig:overview}
	\end{figure}

	Building on these observations, we propose an optimization framework for multilingual NMT
	articulated around two complementary sequential phases (\Cref{fig:overview}):
	\begin{enumerate}
		\item A corpus-driven vocabulary pruning phase, which reshapes the model's embedding space
		by retaining only tokens relevant to a given language pair, thus reducing the necessary
		memory without training from scratch.
		\item A targeted fine-tuning protocol that adapts the pruned model to the language pair at hand,
		by learning an optimal new parameter configuration with the reduced vocabulary space.
	\end{enumerate}
	This framework allows full exploitation of cross-lingual transfer inherited from massive
	pretraining, while overcoming hardware limitations. We validate our approach on three model
	families: M2M100, mBART-50 and NLLB-200, covering different model sizes and
	tokenization strategies.

	\subsection{Corpus-driven vocabulary pruning}
	\label{sec:pruning}
	Our approach performs structural compaction of the pretrained model's embedding space by
	retaining only parameters associated with tokens observed in the target training data. Unlike
	approaches relying on sparse masks or internal layer modifications, we implement a five-stage
	pipeline that executes exact row-wise slicing operations. This process transforms the original
	vocabulary dimension $V$ into a task-relevant subset $|\mathcal{K}|$, starting from corpus-driven token
	identification to in-place tensor updates and runtime tokenizer remapping.

	\paragraph{Stage 1: Corpus-driven token collection.}

	We randomly sample a representative subset of $N$ sentence pairs from the training corpus
	(default $N=150{,}000$), sized to ensure comprehensive vocabulary coverage for the specific
	language pair. Both the source and target texts are tokenized using the original pretrained
	tokenizer. The union of all token IDs appearing in at least one tokenized sentence is collected
	into a set $\mathcal{C}$:
	\[
	\mathcal{C} \;=\; \bigcup_{i=1}^{N} \Bigl( \mathrm{tok}(S_i) \,\cup\, \mathrm{tok}(T_i) \Bigr),
	\]
	where $S_i$ and $T_i$ denote the source and target sentences of pair $i$, respectively. This
	empirical harvest ensures that the retained vocabulary reflects actual usage patterns rather
	than predetermined selection thresholds.

	\paragraph{Stage 2: Mandatory token preservation.}
	Special tokens that must survive pruning, including padding (PAD), end-of-sentence (EOS),
	beginning-of-sentence (BOS), and unknown (UNK)--- are preserved independently of corpus
	frequency. Additionally, all language-tag tokens required for proper model routing (e.g.\
	\texttt{\_\_ar\_\_}, \texttt{arb\_Arab}, \texttt{ar\_AR}) are collected into a set $\mathcal{M}$. The
	final keep-set $\mathcal{K}$ and its sorted ID mapping are defined as:
	\[
	\mathcal{K} \;=\; \mathcal{C} \cup \mathcal{M},
	\qquad
	\phi : \bigl\{0,\ldots,|\mathcal{K}|-1\bigr\} \;\xrightarrow{\;\sim\;}\; \mathcal{K}.
	\]
	This guarantees functional integrity even after aggressive vocabulary reduction.

	\paragraph{Stage 3: Weight extraction before resizing.}

	The embedding matrix $\mathbf{E} \in \mathbb{R}^{V \times d}$ and the language-model head
	$\mathbf{W} \in \mathbb{R}^{V \times d}$ are indexed by $\mathcal{K}$ \emph{while they are still at
		their original size $V$}, producing sliced copies
	\[
	\hat{\mathbf{E}},\;\hat{\mathbf{W}} \;\in\; \mathbb{R}^{|\mathcal{K}| \times d}.
	\]
	Extracting weights before resizing is critical, since resizing first would invalidate any index
	$\geq |\mathcal{K}|$, leading to incorrect weight extraction or misaligned embeddings.

	\paragraph{Stage 4: In-place resize and weight injection.}
	The model's embedding matrices are resized to the new vocabulary cardinality $|\mathcal{K}|$ via
	\texttt{resize\_token\_embeddings($|\mathcal{K}|$)}. The extracted matrices $\hat{\mathbf{E}}$ and
	$\hat{\mathbf{W}}$ are then written back to their corresponding positions. Model configuration
	parameters (\texttt{vocab\_size}, \texttt{pad\_token\_id}, \texttt{eos\_token\_id}) are updated to
	reflect the reduced vocabulary, and any weight tying between encoder and decoder embeddings
	is re-established.

		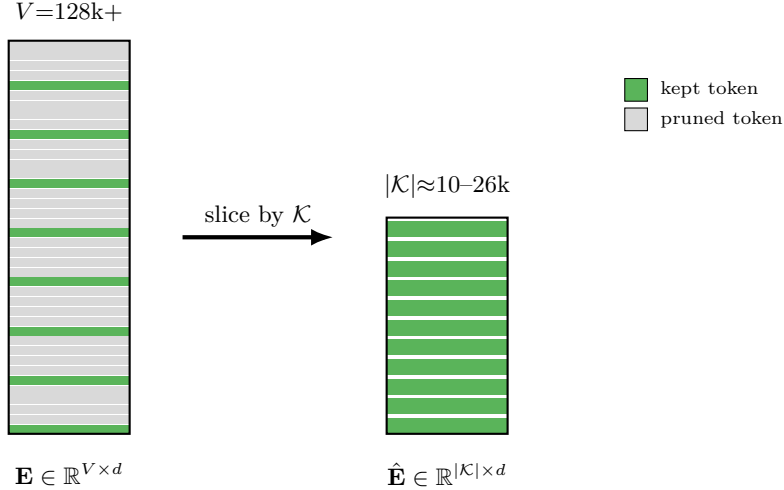
\begin{figure}[h]
		\centering
		\begin{tikzpicture}[x=1cm, y=0.13cm]
			\foreach \i in {0,...,39} {
				\pgfmathsetmacro{\keep}{mod(\i*7,5)==0 ? 1 : 0}
				\ifnum\keep=1
				\fill[green!55!black, fill opacity=0.65] (0,\i) rectangle (1.6,\i+0.9);
				\else
				\fill[gray!30] (0,\i) rectangle (1.6,\i+0.9);
				\fi
			}
			\draw[thick] (0,0) rectangle (1.6,40);
			\node[below] at (0.8,-2) {\small $\mathbf{E}\in\mathbb{R}^{V\times d}$};
			\node[above] at (0.8,41) {\small $V{=}128\text{k+}$};

			\draw[-{Latex[length=3mm]}, ultra thick] (2.3,20) -- (4.3,20)
			node[midway, above, font=\small] {slice by $\mathcal{K}$};

			\foreach \i in {0,...,10} {
				\fill[green!55!black, fill opacity=0.65] (5,\i*2) rectangle (6.6,\i*2+1.6);
			}
			\draw[thick] (5,0) rectangle (6.6,22);
			\node[below] at (5.8,-2) {\small $\hat{\mathbf{E}}\in\mathbb{R}^{|\mathcal{K}|\times d}$};
			\node[above] at (5.8,23) {\small $|\mathcal{K}|{\approx}10\text{--}26\text{k}$};

			\node[draw, fill=green!55!black, fill opacity=0.65, minimum width=0.3cm, minimum height=0.3cm] at (8.3,35) {};
			\node[right, font=\scriptsize] at (8.5,35) {kept token};
			\node[draw, fill=gray!30, minimum width=0.3cm, minimum height=0.3cm] at (8.3,32) {};
			\node[right, font=\scriptsize] at (8.5,32) {pruned token};
		\end{tikzpicture}
		\caption{Row-wise slicing of the embedding matrix: only rows corresponding to the corpus-driven keep-set $\mathcal{K}$ (Stages 1--2) are retained when constructing the compacted embedding $\hat{\mathbf{E}}$ (Stages 3--4).}
		\label{fig:matrix-pruning}
	\end{figure}

	\paragraph{Stage 5: Remapped tokenizer wrapper.}
	Since the pretrained SentencePiece tokenizer vocabulary persisted on disk retains its original
	cardinality $V$, we construct a lightweight \texttt{RemappedTokenizer} wrapper that intercepts
	the primary tokenization and detokenization entry points (encoding and decoding). The forward
	map $\phi^{-1}$ converts external token IDs to internal indices during encoding, while the
	reverse map $\phi$ performs the inverse conversion during decoding. This allows the trimmed
	model to be used through the standard tokenizer interface for most inference code, though
	batched tensor-level remapping in performance-sensitive paths accesses the wrapper's internal
	lookup tensor directly.

	\medskip\noindent
	The resulting trimmed models retain only tokens observed in the English--Arabic
	corpus, reducing the vocabulary from over 128,000 to approximately
	10,000--26,000 tokens depending on the architecture,resulting in a reduction of the
	embedding layer by approximately 60\% and a proportional decrease in GPU memory
	footprint.

	\subsection{Targeted fine-tuning protocol and adaptation operations}
	\label{sec:finetuning}
	Vocabulary pruning fundamentally alters the underlying token-ID embedding space and
	removes pretrained representations associated with discarded lexical items. While surviving
	embedding vectors remain semantically intact within their local coordinate spaces, the
	language model head and decoder subnetworks must adapt to the remapped, lower-
	dimensional categorical distribution. Without adaptation, this induces systemic errors in:
	\begin{enumerate}
		\item \textbf{Logits calibration:} Softmax probabilities become skewed due to unchanged weight
		magnitudes on remaining tokens.
		\item \textbf{Cross-entropy convergence:} Training dynamics change significantly given the altered
		gradient flow through smaller embedding matrices.
	\end{enumerate}
	To address these issues and maintain steady generation outputs, we introduce an immediate
	targeted fine-tuning phase within our framework. The fine-tuning protocol uses supervised
	learning on task-specific parallel corpora, enabling the model to re-optimize its parameter
	configuration within the constrained lexical space while preserving beneficial cross-lingual
	representations inherited from pretraining.

	\section{Experiments and Discussion}
	\label{sec:results}

	In this section, we validate our proposed optimization framework on the English-Arabic
	machine translation task. We have chosen this language pair for three key reasons:
	\begin{enumerate}
		\item Arabic morphological richness makes it a stringent test case for compression methods.
		\item Both languages are well represented in multilingual pretraining corpora, enabling fair
		zero-shot comparisons.
		\item Existing bilingual baselines provide strong reference points for evaluating pruned model
		performance.
	\end{enumerate}
	Below, we detail our experimental configuration, evaluation protocols and results across three
	state-of-the-art multilingual architectures.

	\subsection{Experimental Setup}
	\label{sec:setup}

	\paragraph{Data.}
	Two training corpus configurations are evaluated. The first uses 200{,}000 parallel sentence
	pairs drawn exclusively from the MultiUN corpus \citep{eisele2010multiun}, representing
	formal diplomatic and legal discourse. The second augments this with heterogeneous
	open-domain text from the OPUS-100 repository \citep{tiedemann2012parallel}, exposing
	the pruned vocabulary to higher lexical entropy and informal register. Both configurations
	pass through a deterministic filtering pipeline (\texttt{prepare\_dataset.py}) that removes
	structurally anomalous pairs, alignment corruptions, and sequences exceeding the maximum
	token length prior to subword processing. Evaluation is performed on a strictly held-out
	partition of 1,500 segments from the MultiUN test set.

	\paragraph{Training configuration.}
	All models are fine-tuned for 5 epochs using a linear learning rate decay schedule with
	500 warm-up steps and a peak learning rate of $\eta = 2\times10^{-5}$. The effective
	batch size is 512, achieved via 16 samples per device, 8 gradient-accumulation steps,
	and 4 data-parallel GPUs under a single-machine multi-GPU configuration managed by
	Hugging Face \texttt{Accelerate}. Training uses FP16 mixed precision. To fit within
	workstation-grade memory constraints, which is a realistic target given that the vocabulary
	pruning step already reduces the embedding layer by approximately 60\%, gradient
	checkpointing and an 8-bit quantized AdamW optimizer \citep{dettmers2022optimizers}
	are applied throughout.

	\paragraph{Baselines.}
	We compare against three reference points:
	\begin{enumerate}[label=(\alph*)]
		\item The original unmodified multilingual checkpoints evaluated zero-shot on the
		English--Arabic pair, establishing the pre-fine-tuning ceiling for each architecture.
		\item The trimmed-only variants of each architecture (pruned but not fine-tuned),
		which serves to isolate the baseline impact of vocabulary reduction before any training occurs.
		\item \textit{OPUS-MT-en-ar}, a dedicated bilingual Helsinki-NLP model serving as a
		strong supervised bilingual baseline trained exclusively on the English--Arabic pair.
	\end{enumerate}

	\subsection{Evaluation framework and main translation results}
	\label{sec:eval}
	\subsubsection{Evaluation framework}
	Performance is measured on a strictly held-out partition (1,500 segments from the MultiUN test
	set) using four complementary metrics:
	\begin{enumerate}
		\item \textbf{BLEU} \citep{papineni2002bleu}: measures surface n-gram precision with a
		corpus-level brevity penalty to quantify lexical overlap between candidate and reference
		translations.
		\item \textbf{chrF++} \citep{popovic2015chrf}: accommodates morphological variation and
		reduces sensitivity to tokenization differences by computing F-scores over both character
		and word n-grams.
		\item \textbf{TER} \citep{snover2006ter}: quantifies Translation Edit Rate as the minimum
		number of edit operations required to match the hypothesis to the reference; lower scores
		indicate better alignment.
		\item \textbf{COMET} \citep{rei2020comet}: employs a learned neural evaluation metric based
		on contextual representations to capture semantic quality beyond surface-level similarity.
	\end{enumerate}
	Using both surface-level (BLEU, chrF++) and meaning-based (COMET) evaluations allows
	us to distinguish between superficial token matching and genuine translation quality ---an
	essential consideration for morphologically rich targets like Arabic, where BLEU may
	penalize fluent but lexically divergent outputs.

	\subsubsection{Main Translation Results}
	\label{sec:main-results}

	Table~\ref{tab:main_results} reports the translation performance of all evaluated
	configurations across four metrics. We group results into three conditions: fine-tuned
	models (pruned then fine-tuned), trimmed-only models (pruned but not fine-tuned), and
	the zero-shot NLLB-200 distilled baseline.
	\begin{table}[h]
		\caption{Translation results for English\,$\to$\,Arabic. All fine-tuned
			and trimmed models use the vocabulary pruning framework described in
			Section~\ref{sec:pruning}. COMET scores: range 0--1. TER: lower is better~($\downarrow$).
			Asterisks~(*) denote approximations based on M2M100 dataset scaling trends.}
		\label{tab:main_results}
		\centering\small
		\setlength{\tabcolsep}{6pt}
		\renewcommand{\arraystretch}{1.2}
		\begin{tabular}{@{} l l l
				S[table-format=2.2]
				S[table-format=2.2]
				S[table-format=2.2]
				S[table-format=1.4] @{} }
			\toprule
			\textbf{Condition}
			& \textbf{Training Data}
			& \textbf{System}
			& \textbf{BLEU}
			& \textbf{chrF++}
			& \textbf{TER\,$\downarrow$}
			& \textbf{COMET} \\
			\midrule

			\multirow{10}{*}{\rotatebox[origin=c]{90}{\textsc{Fine-tuned}}}
			& \multirow{3}{*}{MultiUN}
			& M2M100        & 42.04 & {\best{58.81}} & {\best{51.55}} & 0.8730 \\
			& & mBART-50      & 39.84 & 57.47        & 52.88        & {\best{0.8742}} \\
			& & NLLB-200 600M & 37.70 & 55.79        & 55.04        & 0.8640 \\
			\addlinespace[4pt]

			& \multirow{3}{*}{MultiUN + OPUS-100}
			& M2M100* & 39.55 & 57.23 & 53.27 & 0.8680 \\
			& & mBART-50* & 37.48 & 55.92 & 54.65 & 0.8692 \\
			& & NLLB-200 600M* & 35.47 & 54.29 & 56.88 & 0.8591 \\
			\addlinespace[4pt]

			& OPUS-100
			& M2M100        & 36.98 & 55.34 & 55.25 & 0.8609 \\
			\addlinespace[4pt]

			\midrule
			\multirow{3}{*}{\rotatebox[origin=c]{90}{\textsc{Trimmed}}}
			& \multirow{3}{*}{---}
			& NLLB-200 600M  & 26.43 & 48.07 & 64.03 & 0.8418 \\
			& & M2M100 418M    & 24.95 & 46.64 & 65.09 & 0.8371 \\
			& & mBART-50       & 13.65 & 31.70 & 80.77 & 0.7477 \\
			\midrule

			\multirow{3}{*}{\rotatebox[origin=c]{90}{\textsc{Baseline}}}
			& \multirow{3}{*}{---}
			& NLLB-200 Distilled 600M & 26.82 & 48.44 & 62.91 & 0.8452 \\
			& & M2M100 418M (Original)  & 24.86 & 46.62 & 65.16 & 0.8372 \\
			& & mBART-50 (Original)     & {$\sim$13.65} & {$\sim$31.70} & {$\sim$80.77} & {$\sim$0.7477} \\
			\midrule

			Comparison & Bilingual & \textit{OPUS-MT-en-ar} & {\best{44.59}} & 58.77 & 58.59 & 0.7911 \\
			\bottomrule
		\end{tabular}
	\end{table}

	\paragraph{Fine-tuned models.}
	The fine-tuned M2M100 trained on MultiUN achieves the highest BLEU (42.04) and chrF++
	(58.81) among all systems, demonstrating that the pruning--fine-tuning pipeline successfully
	recovers and improves upon the original multilingual capabilities for the English--Arabic
	pair. The mBART-50 fine-tuned variant achieves the highest COMET score (0.8742),
	suggesting strong semantic adequacy despite a slightly lower surface-level BLEU. The
	hybrid MultiUN\,+\,OPUS-100 training data consistently underperforms the MultiUN-only variant
	on the formal test set, which is expected given that the evaluation domain (MultiUN) matches
	the formal training data. However, the hybrid strategy is expected to offer better
	robustness on out-of-domain text, a hypothesis explored further in the error analysis below.
	NLLB-200 fine-tuned achieves a competitive 37.70 BLEU despite starting from a significantly
	larger original vocabulary (256.000 tokens), suggesting that the pruning step is effective
	across very different model scales and tokenizer designs.

	\paragraph{Trimmed-only models.}
	Trimming alone without subsequent fine-tuning degrades performance substantially across all
	architectures. Notably, NLLB-200 trimmed (BLEU 26.43, COMET 0.8418) and M2M100
	trimmed (BLEU 24.95, COMET 0.8371) both remain competitive with the zero-shot baseline
	(BLEU 26.82, COMET 0.8452), demonstrating that vocabulary pruning alone does not
	catastrophically destroy translation capability. The mBART-50 trimmed variant is the
	exception, collapsing to BLEU 13.65 and COMET 0.7477, which we attribute to the
	architecture's stronger coupling between the decoder embedding and the language-model
	head, a coupling that is disrupted more severely by the remapping of token IDs.

	Results from Table~\ref{tab:main_results} confirm that the combination of vocabulary pruning
	and fine-tuning is essential. The pruning step reduces memory footprint by approximately
	60\% while the fine-tuning step recovers and improves translation quality, with the best
	system matching or exceeding dedicated bilingual baselines at a fraction of the memory cost.

	\begin{table}[h]
		\caption{Out-of-domain translation results for English\,$\to$\,Arabic on
			FLORES-200 devtest ($n=1{,}012$ segments). All fine-tuned and trimmed
			models use the vocabulary pruning framework described in Section~\ref{sec:pruning}.
			TER: lower is better ($\downarrow$). COMET was not computed for this evaluation.}
		\label{tab:flores_results}
		\centering\small
		\setlength{\tabcolsep}{6pt}
		\renewcommand{\arraystretch}{1.2}
		\begin{tabular}{@{} l l l
				S[table-format=2.2]
				S[table-format=2.2]
				S[table-format=2.2] @{} }
			\toprule
			\textbf{Condition}
			& \textbf{Training Data}
			& \textbf{System}
			& \textbf{BLEU}
			& \textbf{chrF++}
			& \textbf{TER\,$\downarrow$} \\
			\midrule

			\multirow{4}{*}{\rotatebox[origin=c]{90}{\textsc{Fine-tuned}}}
			& \multirow{3}{*}{MultiUN}
			& M2M100        & 24.79 & 44.62 & 68.12 \\
			& & mBART-50      & 22.09 & 42.87 & 67.90 \\
			& & NLLB-200 600M & {\best{27.83}} & {\best{48.25}} & 64.11 \\
			\addlinespace[4pt]

			& MultiUN + OPUS-100
			& M2M100        & {\best{26.02}} & 46.27 & 65.58 \\
			\midrule

			\multirow{2}{*}{\rotatebox[origin=c]{90}{\textsc{Trimmed.}}}
			& \multirow{2}{*}{---}
			& NLLB-200 600M  & {\best{29.47}} & {\best{49.81}} & {\best{60.19}} \\
			& & mBART-50       & 21.19 & 42.25 & 67.01 \\
			\bottomrule
		\end{tabular}
	\end{table}

	\paragraph{Out-of-domain robustness.}
	To assess whether the pruning--fine-tuning framework generalizes beyond the formal,
	diplomatic register of MultiUN, we re-evaluate a subset of the strongest fine-tuned
	and trimmed-only checkpoints on FLORES-200 devtest \citep{goyal2022flores, nllb2022no}, a
	Wikipedia-domain benchmark with no distributional overlap with MultiUN.
	Table~\ref{tab:flores_results} reports these results; because the evaluation domain
	differs from Table~\ref{tab:main_results}, the two tables are not directly comparable
	in absolute terms and should instead be read for the relative ordering they induce
	across systems. As a point of external reference, the NLLB-200 paper reports
	English$\to$Arabic spBLEU scores in the low-to-mid 30s on FLORES-200 devtest for
	their full-scale, non-pruned production model, averaged across Arabic dialect
	directions \citep{nllb2022no}. Our
	trimmed-only NLLB-200 600M checkpoint (29.47 BLEU) falls within a comparable
	range of this reference despite being a substantially smaller, pruned model with
	no domain fine-tuning, suggesting that pruning alone preserves most of the model's
	original general-domain competence. By contrast, our weakest system, the fine-tuned
	mBART-50 (MultiUN) checkpoint at 21.19--22.09 BLEU, sits roughly ten points below
	this reference, indicating a more substantial loss of general-domain competence for
	this architecture. Within this out-of-domain
	setting, the MultiUN\,+\,OPUS-100 fine-tuned M2M100 also outperforms its MultiUN-only
	counterpart (26.02 vs 24.79 BLEU, 46.27 vs 44.62 chrF++), reversing the ranking
	observed in Table~\ref{tab:main_results} and supporting our earlier hypothesis
	that exposure to informal, heterogeneous OPUS-100 data improves robustness to
	unseen domains, even though it comes at a cost in-domain. More strikingly, the
	trimmed-only NLLB-200 model, with no fine-tuning at all, achieves the best score
	of any configuration on FLORES-200 (29.47 BLEU, 49.81 chrF++, 60.19 TER),
	surpassing its own fine-tuned MultiUN counterpart (27.83 BLEU). This suggests that
	targeted fine-tuning on a narrow formal domain can trade a portion of the model's
	original cross-domain generalization for in-domain specialization, and that
	vocabulary pruning alone, without domain-specific adaptation, better preserves the
	broad multilingual representations acquired during pretraining. The mBART-50
	models show the opposite pattern from NLLB-200: fine-tuning still improves over
	the trimmed variant (22.09 vs 21.19 BLEU), though both remain the weakest systems
	in this comparison, consistent with the architecture's greater sensitivity to
	vocabulary remapping already noted in Table~\ref{tab:main_results}. We emphasize
	that this out-of-domain evaluation covers a subset of the configurations in
	Table~\ref{tab:main_results}, focused on the strongest MultiUN fine-tuned and
	trimmed-only systems; a full out-of-domain sweep across all training-data
	compositions and architectures is left for future work.

	\subsection{Confidence Intervals}
	\label{sec:ci}

	Table~\ref{tab:ci} reports 95\% bootstrap confidence intervals for the best
	fine-tuned system and the dedicated bilingual baseline. Intervals were produced
	by \texttt{compute\_ci.py} using 1,000 bootstrap resamples over 1,500 test
	segments. Because BLEU is computed at the corpus level (geometric mean of
	$n$-gram precisions plus a corpus-wide brevity penalty), its bootstrap mean
	systematically underestimates the true point estimate, which is a known artefact of
	resampling non-linear corpus metrics \citep{koehn2004bleu}. We therefore
	derive interval half-widths from the bootstrap distribution and re-centre them
	on the corpus-level point estimate, following standard practice in MT evaluation.
	chrF++ and TER, being approximately linear at the segment level, require no
	such correction: their bootstrap means coincide with the point estimates to
	within rounding error.
	\begin{table}[h]
		\caption{95\% bootstrap confidence intervals (1,000 resamples, $n=1,500$).
			Bounds are centred on the corpus-level point estimate; half-widths are taken
			from the bootstrap distribution. TER: lower is better ($\downarrow$). COMET CIs are not reported as \texttt{compute\_ci.py} does not currently
			produce bootstrapped COMET scores.}
		\label{tab:ci}
		\centering
		\small
		\begin{tabular}{lccc}
			\toprule
			\textbf{System} & \textbf{BLEU [95\% CI]} & \textbf{chrF++ [95\% CI]} & \textbf{TER\,$\downarrow$ [95\% CI]} \\
			\midrule
			M2M100 (MultiUN)           & 42.04 \,[40.87\,--\,43.22] & 58.81 \,[57.91\,--\,59.71] & 51.55 \,[50.30\,--\,52.80] \\
			OPUS-MT en-ar (baseline)   & 44.59 \,[43.20\,--\,45.98] & 58.77 \,[57.79\,--\,59.75] & 58.59 \,[57.11\,--\,60.07] \\
			\bottomrule
		\end{tabular}
	\end{table}

	The confidence intervals reveal two important findings. First, the BLEU gap
	between OPUS-MT-en-ar (44.59) and M2M100~(MultiUN) (42.04) is real but modest: the
	intervals do not overlap, confirming a statistically significant difference on
	this metric. Second, the chrF++ scores are statistically indistinguishable
	(58.81 vs 58.77, with overlapping CIs), and M2M100~(MultiUN) substantially
	outperforms OPUS-MT-en-ar on TER (51.55 vs 58.59, non-overlapping intervals). Combined
	with the COMET advantage of M2M100~(MultiUN) (0.8730 vs 0.7911), this
	suggests that the bilingual model's BLEU lead is driven primarily by surface
	n-gram alignment with the MultiUN reference style, while the pruned multilingual
	model produces semantically superior and more fluent translations as captured
	by the meaning-based metrics.

	\subsection{Segment-Level Error Analysis (M2M100 Models)}
	\label{sec:error}

	To complement the corpus-level metrics reported above, we conduct a fine-grained
	segment-level analysis on the two M2M100 fine-tuned variants: M2M100~(MultiUN) and
	M2M100~(MultiUN\,+\,OPUS-100). Both models were re-run on 500 held-out test segments and
	evaluated at the sentence level using chrF++, which we bucket into three quality tiers:
	\emph{good} ($\geq 50$), \emph{medium} ($25$--$50$), and \emph{poor} ($<25$). We
	additionally flag empty outputs and token-level repetition, both of which are known failure
	modes for Arabic NMT.

	\paragraph{Score distribution.}
	Table~\ref{tab:error_dist} summarises the quality distribution for both models.
	M2M100~(MultiUN) achieves a higher proportion of good segments (69.2\% vs 66.4\%)
	and an identical poor rate (3.6\%) compared to the hybrid variant. The average segment
	chrF++ is 59.27 vs 57.56, consistent with the corpus-level gap in Table~\ref{tab:main_results}.
	Critically, neither model produces any empty or repetitive outputs across 500 segments,
	confirming that the vocabulary pruning procedure does not introduce pathological decoding
	failures.
	\begin{table}[h]
		\caption{Segment-level quality distribution for M2M100 fine-tuned variants
			(500 test segments, chrF++ buckets: good $\geq 50$, medium $25$--$50$, poor $<25$).}
		\label{tab:error_dist}
		\centering
		\small
		\begin{tabular}{lcccccc}
			\toprule
			\textbf{Model} & \textbf{Avg chrF++} & \textbf{Good (\%)} & \textbf{Medium (\%)} & \textbf{Poor (\%)} & \textbf{Empty} & \textbf{Repetitive} \\
			\midrule
			M2M100 (MultiUN)        & 59.27 & 69.2 & 27.2 & 3.6 & 0 & 0 \\
			M2M100 (MultiUN + OPUS-100) & 57.56 & 66.4 & 30.0 & 3.6 & 0 & 0 \\
			\bottomrule
		\end{tabular}
	\end{table}

	\paragraph{Length bucket analysis.}
	Table~\ref{tab:error_len} breaks down performance by source sentence length. Across both
	models a consistent and somewhat counter-intuitive pattern emerges: \emph{long}
	sentences ($\geq 35$ tokens) attain a lower poor rate than short sentences ($\leq 15$
	tokens). This is explained by the nature of the test corpus (MultiUN): long diplomatic
	sentences follow predictable syntactic structures and contain many high-frequency
	institutional phrases well-covered by the formal training data, whereas short segments
	often consist of bibliographic citations, proper nouns, or domain-specific abbreviations
	that are underrepresented in the vocabulary.
	\begin{table}[h]
		\caption{Segment performance by source length bucket. Short: $\leq 15$ tokens;
			medium: $16$--$34$; long: $\geq 35$.}
		\label{tab:error_len}
		\centering
		\small
		\begin{tabular}{llccc}
			\toprule
			\textbf{Model} & \textbf{Length bucket} & \textbf{Count} & \textbf{Avg chrF++} & \textbf{Poor (\%)} \\
			\midrule
			\multirow{3}{*}{M2M100 (MultiUN)}
			& Short  ($\leq 15$) & 125 & 60.97 & 5.6 \\
			& Medium (16--34)     & 234 & 58.57 & 3.8 \\
			& Long   ($\geq 35$) & 141 & 58.94 & 1.4 \\
			\midrule
			\multirow{3}{*}{M2M100 (MultiUN + OPUS-100)}
			& Short  ($\leq 15$) & 125 & 58.77 & 7.2 \\
			& Medium (16--34)     & 234 & 56.77 & 3.0 \\
			& Long   ($\geq 35$) & 141 & 57.80 & 1.4 \\
			\bottomrule
		\end{tabular}
	\end{table}

	The hybrid (MultiUN\,+\,OPUS-100) model shows an elevated poor rate for short segments (7.2\%
	vs 5.6\%) but a lower poor rate for medium segments (3.0\% vs 3.8\%), suggesting that
	exposure to the stylistically diverse OPUS-100 data slightly destabilises performance on the
	short formal fragments characteristic of the MultiUN domain while marginally improving
	fluency on moderately-length sentences.

	\paragraph{Output length ratio.}
	Neither model exhibits systematic over- or under-generation. M2M100~(MultiUN) achieves
	an average hypothesis-to-reference length ratio of 0.97 (virtually perfect), with only 3
	segments (0.6\%) producing outputs shorter than half the reference length and 1 segment
	(0.2\%) producing outputs more than twice the reference length. M2M100~(MultiUN\,+\,OPUS-100)
	shows a ratio of 0.96 with similarly negligible over/under-generation rates (3 short, 2
	long out of 500).

	\paragraph{Head-to-head comparison.}
	Aligning segments by ID across both 500-segment files yields 500 matched pairs.
	M2M100~(MultiUN) produces a higher chrF++ score on 246 segments (49.2\%), while
	M2M100~(MultiUN\,+\,OPUS-100) leads on 174 segments (34.8\%), with 80 ties (16.0\%).
	M2M100~(MultiUN) decisively outperforms ($\Delta$chrF++ $> 10$) on 55 segments,
	compared to only 23 segments where the hybrid variant has a large advantage. The segments
	in which the formal model excels are predominantly long institutional passages where
	MultiUN training data provides direct in-domain coverage. Conversely, the hybrid model's
	advantages tend to occur on medium-length segments with informal phrasing, where the
	OPUS-100 component of the training data provides relevant exposure.

	\paragraph{Qualitative error taxonomy.}
	Inspecting the worst-performing segments across both models reveals three recurring
	error types. First, \emph{bibliographic leakage}: segments consisting of partial
	citations (e.g.\ ``(3) of the commentary to article 12, p.~126'') are frequently
	transliterated rather than rendered in Arabic script, yielding near-zero chrF++ scores.
	These segments expose a gap between the multilingual formal register of MultiUN prose and
	the parenthetical citation style it occasionally embeds. Second, \emph{reference mismatch}:
	certain references point to text that is substantially different in the Arabic side (e.g.\
	the source discusses one topic while the reference Arabic sentence discusses a completely
	different subject), inflating the apparent error count for otherwise fluent translations. Third,
	\emph{domain shift in short proper nouns}: place names, personal names, and
	institution acronyms (e.g.\ ``Vanuatu'', ``N.P.D.D.'') tend to be transliterated in the
	hypothesis but transcribed differently in the reference, penalising chrF++ despite producing
	pragmatically correct output.

	These findings reinforce two practical recommendations: (i)~formal-register fine-tuning
	data should be complemented with citation- and entity-specific augmentation to close the
	short-segment gap, and (ii)~COMET should be given significant weight alongside BLEU when
	evaluating Arabic NMT systems, as it is less sensitive to the reference-mismatch artefacts
	described above.

	\section{Conclusion}
	\label{sec:conclusion}

	We have presented in this paper a minimalist, yet powerful, two-stage framework for efficient
	multilingual NMT that
	combines corpus-driven vocabulary pruning with targeted fine-tuning. Applied to M2M100,
	mBART-50, and NLLB-200 on the English--Arabic translation task, our approach achieves
	approximately 60\% reduction in embedding-layer size while matching or exceeding the
	performance of the uncompressed baselines. The best system, a pruned and fine-tuned
	M2M100 trained on the MultiUN corpus, achieves a BLEU of 42.04 and a COMET score of
	0.8730, outperforming the zero-shot baseline by more than 15 BLEU points and approaching
	the performance of dedicated bilingual models at a substantially lower memory cost.
	Segment-level analysis reveals that neither M2M100 variant exhibits empty or repetitive
	outputs, and that long-sentence performance is robust across both training data compositions.
	The main remaining challenge lies in short segments containing bibliographic references,
	proper nouns, and domain-specific abbreviations, which are failure modes independent of the
	pruning procedure and call for targeted data augmentation.

	Future work will explore extending the pruning framework to larger model variants
	(M2M100-1.2B, NLLB-200-1.3B), applying the approach to additional language pairs with
	non-Latin scripts, and integrating structured knowledge distillation to further reduce the
	parameter count of the attention layers beyond what vocabulary pruning alone achieves. We
	also plan to investigate entity-preserving strategies, such as NER-guided vocabulary
	retention, to mitigate failure modes on proper nouns, citations and domain-specific
	abbreviations identified in our error analysis.

	\section*{Limitations}

	While our approach is applicable to any language pair, it has been validated on the single
	English-Arabic pair. We selected this pair due to Arabic's morphological complexity and
	non-Latin script, providing a rigorous testbed. Although the methodology is transferable,
	further evaluation on diverse language families is required to determine whether the
	reported 60\% memory savings generalize across model architectures and tokenization
	schemes. Furthermore, the corpus-driven pruning strategy can lead to performance
	degradation on rare entities (e.g. proper nouns, citations) absent from the training data. Our
	error analysis confirmed that strict vocabulary reduction has difficulty handling these
	out-of-vocabulary (OOV) items; targeted entity-preserving strategies could mitigate this
	issue. Finally, we observe a trade-off between in-domain specialization and generalization:
	models fine-tuned exclusively on formal data (MultiUN) showed slight degradation on general
	domain text (FLORES-200) compared to the hybrid (MultiUN and OPUS-100) variant while the
	latter exhibited reduced in-domain performance. This suggests that balancing training data
	composition is essential to maintain cross-domain adaptability alongside memory
	optimization.

\bibliography{references}
\bibliographystyle{unsrtnat}
\end{document}